\documentclass{article}
\usepackage[final]{neurips_2024}

\usepackage[normalem]{ulem}
\usepackage{graphicx}
\usepackage[utf8]{inputenc} 
\usepackage[T1]{fontenc}    
\usepackage{hyperref}       
\usepackage{url}            
\usepackage{booktabs}       
\usepackage{amsmath,amssymb,amsfonts,amsthm,bbm}
\usepackage{mathtools}   
\usepackage{stmaryrd} 
\usepackage{enumitem}   
\usepackage{nicefrac}       
\usepackage{microtype}      
\usepackage{xcolor}         
\usepackage{mathrsfs}
\usepackage{comment}
\usepackage{subfigure}
\usepackage{caption}
\usepackage{natbib}
\setcitestyle{numbers,square, sort, compress}

\usepackage{stmaryrd} 

\renewcommand{\hat}{\widehat}

\newcommand{\bfm}[1]{\ensuremath{\boldsymbol{#1}}} 

   \def\bM{\bfm M}

     \def\RR{\mathbb{R}}

   \def\bU{\bfm U}  
     
\def\bw{\bfm w}     
   \def\bX{\bfm X}

 \def\cC{{\cal  C}}
 
 \def\cE{{\cal  E}}

 \def\cT{{\cal  T}}

 \def\cW{{\cal  W}}
 \def\cX{{\cal  X}}

\providecommand{\angles}[1]{\left\langle #1 \right\rangle}
\providecommand{\paren}[1]{\left( #1 \right)}

\usepackage{mathtools}
\DeclarePairedDelimiterX{\infdivx}[2]{(}{)}{%
  #1 \; \delimsize\| \; #2%
}

\newtheorem{definition}{Definition}[section]

\newtheorem{lemma}[definition]{Lemma}

\newtheorem{theorem}[definition]{Theorem}

\definecolor{royalpurple}{rgb}{0.47, 0.32, 0.66}
\definecolor{greenfresh}{HTML}{00897B}
\definecolor{bluefresh}{HTML}{1E88E5}
\definecolor{redfresh}{HTML}{E53935}

\definecolor{royalpurple}{rgb}{0.47, 0.32, 0.66}

\def\beq{\begin{equation}}
\def\eeq{\end{equation}}

\def\bet{\begin{theorem}}
\def\eet{\end{theorem}}

\def\bel{\begin{lemma}}
\def\eel{\end{lemma}}

\usepackage{setspace}
\usepackage{etoolbox}
\BeforeBeginEnvironment{equation*}{\begin{singlespace}\vspace*{-\baselineskip}}
	\AfterEndEnvironment{equation*}{\end{singlespace}\noindent\ignorespaces}
\BeforeBeginEnvironment{equation}{\begin{singlespace}\vspace*{-\baselineskip}}
	\AfterEndEnvironment{equation}{\end{singlespace}\noindent\ignorespaces}
\BeforeBeginEnvironment{align}{\begin{singlespace}\vspace*{-\baselineskip}}
	\AfterEndEnvironment{align}{\end{singlespace}\noindent\ignorespaces}
\BeforeBeginEnvironment{align*}{\begin{singlespace}\vspace*{-\baselineskip}}
	\AfterEndEnvironment{align*}{\end{singlespace}\noindent\ignorespaces}
\BeforeBeginEnvironment{eqnarray}{\begin{singlespace}\vspace*{-\baselineskip}}
	\AfterEndEnvironment{eqnarray}{\end{singlespace}\noindent\ignorespaces}
\BeforeBeginEnvironment{eqnarray*}{\begin{singlespace}\vspace*{-\baselineskip}}
	\AfterEndEnvironment{eqnarray*}{\end{singlespace}\noindent\ignorespaces}

\def\TITLE{TEAFormers: TEnsor-Augmented Transformers for Multi-Dimensional Time Series Forecasting}

\title{\TITLE}
\author{%
Linghang Kong \\
New York University\\
New York, NY 10012 \\
\And
Elynn Chen \\
New York University \\
New York, NY 10012 \\
\AND
Yuzhou Chen \\
University of California, Riverside \\
Riverside, CA 92521 \\
\And
Yuefeng Han \\
Notre Dame University \\
Notre Dame, IN 46556 \\
}

\begin{document}

\maketitle

\begin{abstract}
Multi-dimensional time series data, such as matrix and tensor-variate time series, are increasingly prevalent in fields such as economics, finance, and climate science. Traditional Transformer models, though adept with sequential data, do not effectively preserve these multi-dimensional structures, as their internal operations in effect flatten multi-dimensional observations into vectors, thereby losing critical multi-dimensional relationships and patterns. To address this, we introduce the Tensor-Augmented Transformer (TEAFormer), a novel method that incorporates tensor expansion and compression within the Transformer framework to maintain and leverage the inherent multi-dimensional structures, thus reducing computational costs and improving prediction accuracy. The core feature of the TEAFormer, the Tensor-Augmentation (TEA) module, utilizes tensor expansion to enhance multi-view feature learning and tensor compression for efficient information aggregation and reduced computational load. The TEA module is not just a specific model architecture but a versatile component that is highly compatible with the attention mechanism and the encoder-decoder structure of Transformers, making it adaptable to existing Transformer architectures. Our comprehensive experiments, which integrate the TEA module into three popular time series Transformer models across three real-world benchmarks, show significant performance enhancements, highlighting the potential of TEAFormers for cutting-edge time series forecasting.
\end{abstract}

\section{Introduction}

In the era of big data, multi-dimensional time series data, such as matrix and tensor-valued time series, are increasingly common in applications like economics, finance, and climate science. For instance, policymakers track quarterly economic indicators like GDP growth and inflation across multiple countries \citep{chen2020constrained}; investors monitor financial metrics such as asset/equity ratios and revenue from various companies \cite{chen2023statistical}; and scientists observe hourly environmental variables like PM2.5 and ozone levels at multiple stations \cite{chen2020modeling,chen2024semi}. All these datasets naturally present themselves as time series of matrices (order-2 tensors).

Transformers \cite{vaswani2023attention} have demonstrated substantial promise in modeling sequential data such as language\cite{zhang2018improving}, text\cite{guo2022longt5}, and time series \cite{wen2023transformers}. However, existing Transformer models fail to preserve the tensor structure of data. The calculations within these architectures effectively flatten multi-dimensional observations into vectors, losing the inherent multi-dimensional relationships and patterns.

This paper fill this gap by introducing the Tensor-Augmented Transformer (TEAFormer), a novel approach designed to preserve and leverage multi-dimensional tensor structures. Multiple benefits are achieved by TEAFormer: \textbf{1)} Its {\em tensor expansion} step effectively aggregate all possible useful information by multi-view feature learning from multiple channels, including those constructed from multi-head attention mechanism; \textbf{2)} Its {\em tensor compression} step effectively compress information to a more efficient feature by automatic tensor decomposition, reduce the computational burden associated in the tensor expansion step such as self-attention, and simultaneously enhance their performance. By computing self-attention on this reduced-size core tensor, we achieve significant computational savings without sacrificing the model's ability to capture intricate tensor structures and dependencies in the data; \textbf{3)} TEAFormer is not merely a specific model architecture. Its tensor expansion and compression steps are highly compatible with attention mechanism and the encoder-decoder structure. Thus, we generalize it as \textbf{a TEA-module that can be incorporated in any Transformer model}. 

We incorporate the Tensor-Augmentation module into three widely used time series Transformer models: Transformer \cite{vaswani2023attention}, Informer \cite{haoyietal-informer-2021}, and Autoformer \cite{wu2022autoformer}. We conduct extensive experiments across a diverse range of datasets, demonstrating that our approach not only reduces computational costs but also enhances prediction accuracy. Our results show substantial improvements in evaluation metrics, such as Mean Absolute Error (MAE) and Mean Squared Error (MSE) scores, compared to baseline models. This underscores the potential of tensor decomposition as a powerful tool for advancing matrix and tensor time series forecasting.

Recent statistical studies have explored the benefits of maintaining the multi-dimensional structure in tensor time series using linear factor models \cite{wang2019,chen2020constrained,chen2023statistical, chen2024semi,chen2022factor,han2020iterative,han2022rank, han2023tensor,han2024cp} and linear vector auto-regressive (VAR) models \cite{chen2021autoregressive,han2023rr,li2021multi}. They have shown great advantages of preserving the multi-dimensional structure in tensor time series. However, all of them adopt linear approaches and  have limited capacity to capture potential nonlinear relationships, especially in applications involving language and text.

This work is the first to formally introduce a TEnsor-Augmented Transformer (TEAFormer), distinct from statistical VAR models. Maintaining tensor structure in Transformers is much more challenging due to the architecture's complexity, including encoding, decoding, and multi-head attention mechanisms, as detailed in Section \ref{tea_transformer}. Our contributions are summarized in three key points:
\begin{itemize}
\item \underline{\it Introduction of TEAFormer.} We introduce TEAFormer, a novel multi-dimensional (tensor) time series Transformer architecture that integrates a Tensor-Augmentation module with any Transformer-based models. This is the first work to utilize tensor learning in Transformer.
\item \underline{\it Development of Tensor-Augmentation Module.} We develop a Tensor-Augmentation module that involves tensor expansion and compression, effectively aggregating information through multi-view feature learning while reducing the computational burden associated with self-attention and information aggregation.
\item \underline{\it Comprehensive Empirical Evaluation of Appliction to Existing Tranformers.} Our extensive experimental results show that all implemented TEAFormers outperform their baseline counterparts in 34 out of 42 trials across three real-world benchmarks. Ablation studies reveal that TEAFormer holds significant potential for future research: (1) the tensor compression structure can be further optimized; (2) it excels in learning from small datasets with shorter sequences; (3) tensor compression enhances performance when tensor expansion does not sufficiently aggregate information for learning.
\end{itemize}

The rest of this paper is organized as follows: Section~\ref{related_work} shows related work on Neural Networks and Transformer models for time series forecasting. Section~\ref{tea_transformer} provides a detailed description of the TEAFormer architecture and its integration of tensor decomposition. In Section~\ref{experiment_sec}, we present our experimental setup, including datasets, evaluation metrics, and implementation details, followed by a discussion of the results and an analysis of performance improvements. Finally, Section~\ref{con_future} concludes the paper and outlines potential directions for future research.

\subsection{Related Work}\label{related_work}

Our research lies at the intersection of tensor learning, multidimensional time series analysis, and transformer-based forecasting. Given the extensive literature across these fields, we focus our survey on the most relevant works.

\noindent
\textbf{Multi-dimensional Time Series Analysis.}
Matrix- and tensor-variate time series analysis has emerged as a rapidly growing field in statistical learning, with substantial developments in recent years.
The fundamental challenge in multi-dimensional time series analysis is understanding the intricate relationships among multiple time-dependent variables.
Traditional approaches for handling high-dimensionality and inter-variable correlations involve vectorizing tensor time series into vector time series and applying factor models.
Factor models have emerged as a crucial tool for capturing common dependencies among multiple covariates \cite{bai2002determining,bai2003inferential,bai2013panel,chen2024time}.
Recent research has extended factor model methodologies to matrix- and tensor-variate time series analysis. These studies predominantly employ Tucker low-rank structures \citep{chen2023statistical,chen2023community,chen2020constrained,chen2024semi,chen2022analysis,chen2022factor,han2020iterative,han2022rank,yu2024dynamic,zhou2024factor} and Canonical Polyadic (CP) low-rank structures \citep{chang2023modelling, chen2024estimation, han2024cp, han2023tensor}. Additionally, researchers have developed matrix- and tensor-variate autoregressive models with bilinear or multilinear forms, supported by theoretical guarantees demonstrating that preserving multi-dimensional structure enhances performance \cite{chen2021autoregressive,Hoff2015,li2021multi,li2024cointegrated,han2023rr,chen2020modeling,liu2022identification,chen2022modeling,chen2024factor,han2024simultaneous}.
The computer science literature focuses primarily on empirical approaches, leveraging neural network architectures for forecasting. Variables exhibit distinct patterns and cycles, interacting through both lagged relationships and simultaneous influences. While CNNs and RNNs have been traditional tools, recent innovations include models like DeepGLO \cite{sen2019think}, which employs multiple Temporal Convolutional Networks (TCNs) to capture global and local dependencies. Graph Neural Networks (GNNs) represent another promising direction: STGCL \cite{liu2022when} utilizes contrastive learning at both graph and node levels for spatiotemporal GNNs, while GNN-RNN \cite{fan2022gnnrnn} processes diverse data types through CNNs and RNNs before using GraphSage GNN to extract geospatial representations. CausalGNN \cite{CausalGNN2022} combines various embeddings to obtain latent matrix representations, integrating them with causal embeddings through GNN-based non-linear transformations for disease forecasting.
Despite these advances, the incorporation of tensor structures, particularly Tucker or CP low-rank constraints, into transformer architectures remains largely unexplored, presenting an important gap in the current literature.

\noindent
\textbf{Transformer for Multivariate Time Series Forecasting.}
Transformers \cite{vaswani2023attention} offer significant advantages for handling sequence data, particularly due to their attention mechanisms, which allow for greater parallelization and faster training times on modern hardware. Informer \cite{haoyietal-informer-2021} modifies the traditional transformer architecture for time series forecasting by using a \textit{probsparse} attention mechanism for improved computational efficiency and a generative decoder to reduce complexity and prevent cumulative errors.
Building on Informer, Spacetimeformer \cite{grigsby2021longrange} incorporates spatial correlations by adding time and space information during embedding, calculating both global and local attention. Autoformer \cite{wu2022autoformer} enhances efficiency and decomposition ability through an autocorrelation mechanism and time series decomposition methods. Fedformer \cite{zhou2022fedformer} leverages signal processing techniques, using Fourier transformation for global information decomposition. ETSformer \cite{woo2022etsformer} combines traditional time series decomposition with exponential smoothing, creating interpretable and decomposable forecasting outputs.
Crossformer \cite{zhang2022crossformer} introduces a two-stage attention mechanism, computing attention along the time axis first and then along the dimension axis, with an innovative router mechanism to manage attention distribution and reduce time complexity. DSformer \cite{yu2023dsformer} collects global and local information through sampling methods, rearranging data along the time and dimension axes.

\noindent
\textbf{Tensor-Augmented Neural Networks.}
While neural networks can effectively process high-dimensional tensor inputs, most existing approaches utilize tensors primarily for computational efficiency rather than leveraging their statistical properties. \cite{cohen2016expressive} established a foundational connection between deep networks and hierarchical tensor factorizations. Building on this, tensor contraction layers \citep{kossaifi2017tcl} and regression layers \citep{kossaifi2020trl} were developed to regularize models, successfully reducing parameter counts while preserving model accuracy. Recent innovations, including the Graph Tensor Network \citep{xu2023graph} and Tensor-view Topological Graph Neural Network \citep{wen2024tensor}, have introduced novel frameworks for processing large-scale, multi-dimensional data. Additionally, significant progress has been made in uncertainty quantification and representation learning for tensor-augmented neural networks \cite{chen2024high09,chen2024high10,wu2024tensor,wu2024cproc,wu2024conditional}. Despite these advances, the development of tensor-augmented transformers specifically designed for multi-dimensional time series remains an open challenge.

\section{Tensor-Augmented Transformer}\label{tea_transformer}

For a clear presentation, we consider the most common case of a two-dimensional matrix-variate time series $\bX_t \in \mathbb{R}^{D_1 \times D_2}$ for $t \in [L]$. 
In sequential forecasting, one aims to predict the future values $\bX_{t+1}$ based on historical observations $\bX_1,...,\bX_t$. 

\begin{figure}[ht!] 
\centering
\includegraphics[width=0.8\textwidth]{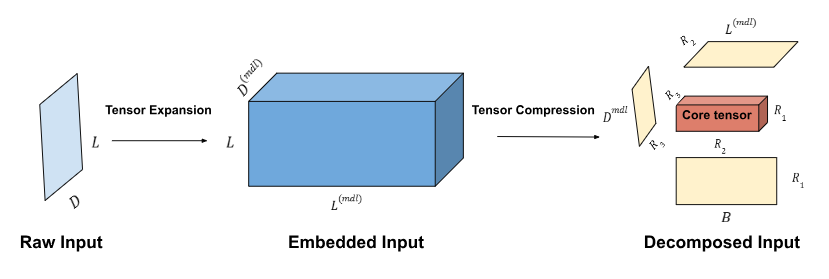}
\caption{The structure of tensor-augmentation module when $E=1$ and $M=1$.\label{tea_module}}
\end{figure}


We preserve the inherent multi-dimensional structure in the TEAFormer. For matrix-variate time series, we let $\cX^{(raw)} \in \mathbb{R}^{L\times D_1\times D_2}$ be one sample of raw input data, where $L$ denotes the sequence length, $D_1$ and $D_2$ denote the sizes of two-dimensional matrix observation. 

\noindent{\bf{Embedding with Tensor Mode and Size Expansion.}} 
During embedding, we transform the raw input $\cX^{(raw)}$ to feature $\cX \in \mathbb{R}^{L\times L^{(mdl)}\times D^{(mdl)}}$. Here, the temporal embedded dimension $L^{(mdl)}$ is not restricted to a scalar value of one-dimension. Instead, it can represent multiple dimensions $L^{(mdl)} = L^{(mdl)}_1\times\cdots\times L^{(mdl)}_E$, where $E\ge 1$ is the total number of modes extended from the temporal mode of $\cX^{(raw)}$. For example, CrossFormer \cite{zhang2022crossformer} expands the raw temporal mode of length $L$ to $L^{(mdl)}= L^{(mdl)}_1 \times L^{(mdl)}_2$ where $L^{(mdl)}_1$ represents the number of sub-sequences and $L^{(mdl)}_2$ represents the embedding of each sub-sequence. This corresponding to {\em Tensor mode expansion}. 

Analogously, the feature embedded dimensions $D^{(mdl)} := D^{(mdl)}_1 \times\cdots\times D^{(mdl)}_M$ can be multi-dimensional with $M \ge 2$. If $M > 2$ where $2$ is the number of dimensions of the raw time series observation $\bX_t$, the embedding results in a {\em mode-expanded feature space}, or {\em Tensor mode expansion}, that is the number of modes increases. If $\Pi_{m=1}^M D^{(mdl)}_m > \Pi_{m=1}^2 D_m$, this embedding results in a {\em size-expanded feature space}, or {\em Tensor size expansion}, that is the size of the feature space increases.
For example, CrossFormer \cite{zhang2022crossformer} expands the raw temporal mode of length $L$ to $L^{(mdl)}= L^{(mdl)}_1 \times L^{(mdl)}_2$ by segmenting the $L$-length sequence to $S$-number of sub-sequences of size $L/S$. The embedding also results in a {\em size-expanded feature space} since $L^{(mdl)}_1 = S$ and $L^{(mdl)}_2 > L/S$. Finally, we note that this setting also include the classic transformer as sub-cases. Specifically, for vector (order-1 tensor) data, $D^{(mdl)}:=D^{(mdl)}_1$ corresponds to the common transformer; for matrix (order-2 tensor) data $D^{(mdl)} := D^{(mdl)}_1 \times D^{(mdl)}_2$ corresponding to matrix time series without {\em Tensor mode expansion}. 

\noindent{\bf{Tensor-Augmented Sequence-to-Sequence Multi-Head Attention (MHA).}}  We first characterize the formulation of a Single-Head Attention (SHA). The embedded feature is $\cX \in \mathbb{R}^{L\times L^{(mdl)}\times D^{(mdl)}}$.
One attention head is structured as a tensor with dimension expressed as $D^{(attn)} := D^{(attn)}_1 \times\dots\times D^{(attn)}_M$. 
The value weight tensor is $\cW_V \in \mathbb{R}^{L^{(mdl)}\times D^{(mdl)}\times D^{(attn)}}$, the query and key weight tensors are $\cW_Q,\cW_K \in \mathbb{R}^{L^{(mdl)}\times D^{(mdl)} \times D^{(attn)}}$, and the output weight tensor is $\cW_O \in \mathbb{R}^{D^{(attn)}\times L^{(mdl)}\times D^{(mdl)}}$, all of which are trainable. The output of one layer of the Transformer is a $(L\times L^{(mdl)}\times D^{(mdl)})$-dimensional tensor, and can be described as follows:
\begin{equation} \label{eqn:one-layer-sha}
\cT^{\rm SHA}(\cX;\cW_Q,\cW_K,\cW_V,\cW_O) := \sigma\paren{{\rm RowSoftmax}\paren{ \angles{\angles{\cX,\cW_Q},\;\angles{\cX,\cW_K}^\top}}\angles{\cX,\cW_V}} \cW_O,
\end{equation}
where $\angles{\cX,\cW_Q}$, $\angles{\cX,\cW_K}$, $\angles{\cX,\cW_V}$ denote tensor inner products that result in a $(L\times D^{(attn)})$-dimensional matrix,  $[{\rm RowSoftmax}(\bM)]_{\ell,:} := {\rm softmax}(\bM_{\ell,:})$ that runs {\rm softmax} on row of its input, and $\sigma$ is a $L_\sigma$ Lipshitz activation function that is applied element-wise and has the property $\sigma(0)=0$.

For MHA, let $H$ denote the number of head, we collect all weight tensors in $\cW := \{\cW_{Q,h},\cW_{K,h},\cW_{V,,h},\cW_{O,h}\}_{h=1}^H$ for all heads $h\in[H]$ and define an extra head weight vector $\bw_H$ of dimension $H$. One layer of the Transformer with MHA can be described as $\cT$
\begin{equation} \label{eqn:one-layer-transformer}
\cT(\cX;\cW, \bw_H) = [\cT^{\rm SHA}(\cX;\cW_{Q,h},\cW_{K,h},\cW_{V,,h},\cW_{O,h})]_{:::h}\times_{4} \bw_H. 
\end{equation}
The dimension of $\cT(\cX)$ is $L\times L^{(mdl)}\times D^{(mdl)}$. Thus, a $S$-multi-layer Transformer can be constructed by iteratively compose $\cT(\cX)$ for $S$ times, denoted as
\begin{equation} \label{eqn:multi-layer-transformer}
\cT_S(\cX;\{\cW^{(s)},\bw_H^{(s)}\}_{s=0}^S) := \cT^{(S)}\circ\cdots\circ\cT^{(0)}(\cX),
\end{equation}
where $\cT^{(s)} = (\cdot;\cW^{(s)},\bw_H^{(s)})$. 

In our study, multi-head self-attention (MSA) is used. MSA is a specific type of MHA, in which queries, keys, and values all come from the same sequence, i.e. $\cW_Q = \cW_K = \cW_V$, thus allowing the model to capture dependencies within that sequence.

\noindent{\bf{Automated Information Aggregation and Compression.}} The initial feature input and the hidden throughputs are all of order-$(1+E+M)$ tensor structure of dimension $L\times L^{(mdl)}\times D^{(mdl)}$ where $E$ is the number of modes of the hidden temporal embedding and $M$ is the number of modes of the feature model embedding. Current transformers treat all dimensions equally and carry out calculation by flattening all the tensors. The idea of this paper is to preserve the tensor (multi-dimensional) structure. As such, we are able to automate information aggregation and information compression through tensor expansion and compression. 

{\em Tensor compression} refers to the procedure that incorporates low-rank tensor decomposition in the procedure, which can only be achieved when we keep the tensor structure in \eqref{eqn:one-layer-transformer}. We incorporate tensor low-rank structures such as {\it CP low-rank}, {\it Tucker low-rank}, and {\it Tensor Train low-rank}.

Tucker low-rank structure is defined by
\begin{equation} \label{eqn:tucker}
\cX = \cC\times_1 \bU_1\times_2\cdots\times_M \bU_M + \cE,
\end{equation}
where 
$\cE\in\RR^{D^{(mdl)}_1\times\cdots\times D^{(mdl)}_M}$ is the tensor of the idiosyncratic component (or noise) and $\cC$ is the latent core tensor representing the true low-rank feature tensors and $\bU_m$, $m\in[M]$, are the loading matrices.  

The complete definitions of three low-rank structures are given in Appendix~\ref{sec:tensor}. 
CP low-rank \eqref{eqn:cp} is a special case where the core tensor $\cC$ has the same dimensions over all modes, that is $R_m = R$ for all $m\in[M]$, and is super-diagonal. 
TT low-rank is a different kind of low-rank structure, which inherits advantages from both CP and Tucker decomposition.
Specifically, TT decomposition can compress tensors as significantly as CP decomposition, while its calculation is as stable as Tucker decomposition.

\noindent{\bf{Practical Experiment Implementation.}} To our knowledge, there does not exist any Transformer-based time series forecasting model that has more than one hidden dimension in the $D^{(mdl)}$ set. In addition, even though different models have different embedding methods, they are either expanding based upon sequence length or features, and would always result in a three-dimensional embedding. Hence, though it is theoretically possible to have tensor-augmented multi-head attention for higher dimensions, we are only able to test our theories under circumstances where there is only one hidden dimension and the embedded data is in three dimension.

Before passing into the attention block, the embedded input as a three-dimensional tensor will be passed into a Tensor-Augmentation module to be decomposed using Tucker decomposition method and be transformed into a group of factorized tensors.
\begin{equation}
\begin{aligned}
\cX &= Embed(\cX^{(raw)}),\\
\cX &= \cC\times_1 \bU_1\times_2\bU_2\times_3\bU_3,
\end{aligned}
\end{equation}
where $\cC \in \mathbb{R}^{R_1\times R_2\times R_3}, R_1\ll L, R_2\ll L^{(mdl)}, R_3\ll D^{(mdl)}$ is a core tensor that acts as a framework to capture the complex interactions between cross-dimensional and temporal features across the dataset, thus could be leveraged as a latent representation for global information in MTS. $\bU_1 \in \mathbb{R}^{L\times R_1}$, $\bU_2\in \mathbb{R}^{L^{(mdl)}\times R_2}$, $\bU_3\in \mathbb{R}^{D^{(mdl)}\times R_3}$ are factor matrices. They can extract the most predominant features in each dimension while maintaining the original dimension's structure.

As we intend to reduce the computational cost of self-attention while capture the global information about the interactions within the data, we will pass the core tensor, which is much smaller in size comparing to the embedded input while preserving essential information, into the attention block to conduct computation. To be more specific, in each encoder layer, we have,
\begin{equation}
\begin{aligned}
\hat{\cC} &= LayerNorm(\cC+MSA(\cC,\cC,\cC)),\\
\hat{\cX}^{enc} &= \hat{\cC}\times_{1}\bU_1\times_{2}\bU_2\times_{3}\bU_3,\\
\end{aligned}
\end{equation}
where $\hat{\cX}^{enc}$ is the output of the current encoder layer. And the rest of the model structure will be the widely adopted Transformer encoder-decoder architecture.
\begin{figure}[ht!] 
\centering
\includegraphics[width=0.7\textwidth]{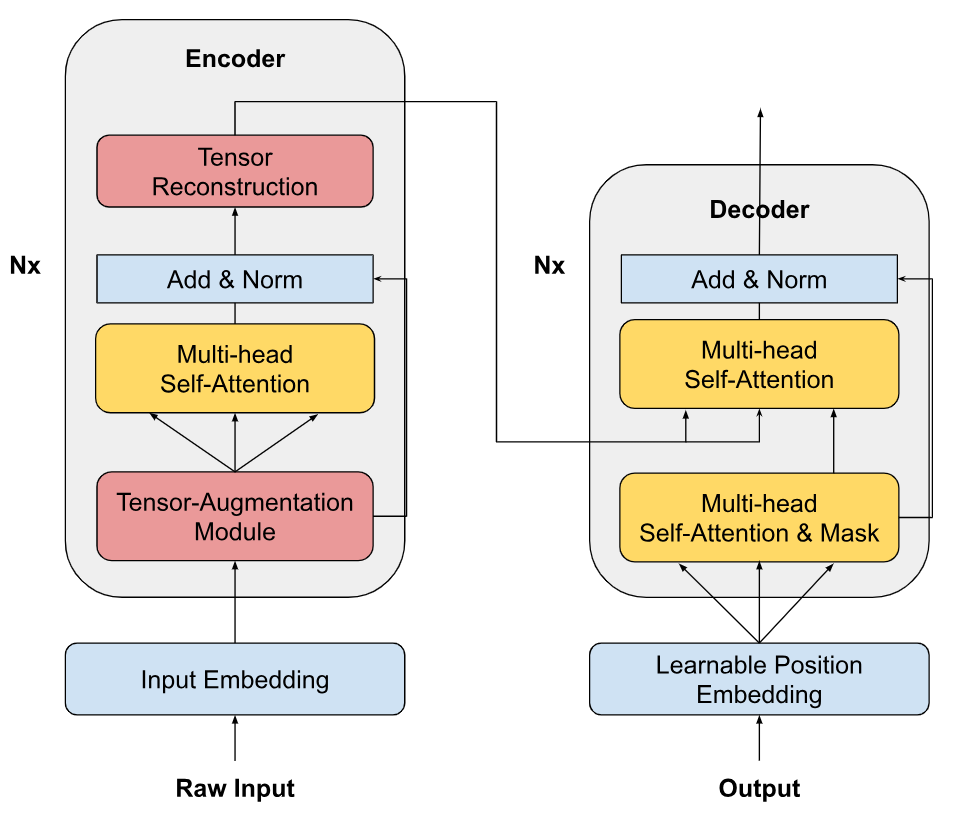}
\caption{The architecture of TEAFormer. The input embedding is a general reference to any positional, segment, or temporal embedding that transforms raw input into 3D shape.\label{teaformer}}
\end{figure}

\vspace{-2ex}
\paragraph{Current Architecture Challenge:}
We assumed that TEAFormer would be much more optimized if we solely pass the core tensor through all the encoders and decoders. However, decoders require a learnable position embedding $\cX^{dec}$ to make the final forecast. The core tensor that is passed as input may mismatch the dimension of $\cX^{dec}$ during MSA computation. Furthermore, we also considered decompose $\cX^{dec}$ to obtain a $\cC^{dec}$ to use as a core tensor for $\cX^{dec}$ in order to solve the dimension mismatch challenge. However, $\cX^{dec}$ is crucial for forecasting, and we suspect that the deducted dimension of $\cC^{dec}$ may not be able to hold as much information as $\cX^{dec}$. Thus we did not adopt this approach and the challenge remains.

\section{Experiments}\label{experiment_sec}
\noindent{\bf Datasets}
We conduct experiments on three real world datasets that are very popular in recent multi-dimensional time series forecasting studies: (1) {ETTh1}: The temperature of electricity transformers (ETT)\footnote{https://github.com/zhouhaoyi/ETDataset} is a vital metric in the long-term deployment of electric power. This dataset contains two years of data from two different counties in China. The ETTh1 subset is the data on 1-hour-level granularity; (2) {ETTm1}: The subset of ETT dataset on 15-minutes-level granularity; and (3) {WTH}\footnote{https://www.ncei.noaa.gov/data/local-climatological-data/}: This dataset includes local climatological data for almost 1,600 locations across the United States, spanning four years from 2010 to 2013. Data points are collected hourly, each consisting of the target value "wet bulb" and 11 additional climate features. The data is split into training, validation, and testing sets with durations of 28, 10, and 10 months, respectively.

\noindent{\bf Baselines}
In this paper, we use three popular attention-based encoder-decoder structure multi-dimensional time series forecasting models as our baseline: (1) {Transformer}~\cite{vaswani2023attention}, 
(2) {Informer}~\cite{haoyietal-informer-2021}, 
and (3) {Autoformer}~\cite{wu2022autoformer}. 

\noindent{\bf Experiment Setup}
To better demonstrate our tensor-augmented attention module's performance, we directly utilized the hyperparameter settings in the baseline papers so that we are able to make comparisons on the performance of the baseline model's finest state. In Autoformer~\cite{wu2022autoformer}, Informer was used as a baseline to be compared with, but the hyperparameters were tuned significantly different from the original Informer~\cite{haoyietal-informer-2021}. In this case, we use the hyperparameter settings from the Informer paper. The prediction windows size $L_y$ is set as follow: {1d, 2d, 7d, 14d, 30d} for ETTh1 and ETTm1, and {4d, 8d, 14d, 30d} for Weather.

\noindent{\bf Evaluation Metrics}
We use two evaluation matrics: Mean Square Error and Mean Absolute Error,
\begin{align*}
{\rm MSE}&= \frac{1}{L L^{(mdl)} D^{(mdl)}} \sum_{t=1}^L \sum_{i=1}^{L^{(mdl)}}\sum_{j=1}^{D^{(mdl)}}\|\bX_{t,ij}-\hat{\bX}_{t,ij}\|_{2}^2 ,\\
{\rm MAE}&= \frac{1}{L L^{(mdl)} D^{(mdl)}} \sum_{t=1}^L \sum_{i=1}^{L^{(mdl)}}\sum_{j=1}^{D^{(mdl)}} |\bX_{t,ij}-\hat{\bX}_{t,ij}|.
\end{align*}

\subsection{Experiment Results and Analysis}
\begin{table}[h]  
\caption{Performance comparison between baselines and TEA-based models.} 
\centering 
\resizebox{\textwidth}{!}{
\begin{tabular}{c c c c c c c c c c c c c c} 
\hline\hline   
Dataset & SeqLen & \multicolumn{2}{c}{Informer} & \multicolumn{2}{c}{TEA-Informer} & \multicolumn{2}{c}{Autoformer} & \multicolumn{2}{c}{TEA-Autoformer} & \multicolumn{2}{c}{Transformer} & \multicolumn{2}{c}{TEA-Transformer}
\\ [0.5ex]  
\hline
& & MSE & MAE & MSE & MAE & MSE & MAE & MSE & MAE & MSE & MAE & MSE & MAE \\
\hline\hline
ETTh1 & 24 & 0.577 & 0.549& \textbf{0.500} & \textbf{0.517} & 0.384 & 0.425 & \textbf{0.372} & \textbf{0.412} & 0.592 & 0.572 & \textbf{0.587} & \textbf{0.545} \\
& 48 & 0.685 & 0.625 & \textbf{0.589} & \textbf{0.578} & 0.392 & \textbf{0.419} & \textbf{0.388} & 0.431 & 0.854 & 0.721 & \textbf{0.784} & \textbf{0.660} \\
& 168 & \textbf{0.931} & \textbf{0.752} & 0.968 & 0.775 & 0.490 & 0.481 & \textbf{0.480} & \textbf{0.479} & 1.047 & 0.833 & \textbf{0.944} & \textbf{0.789} \\ 
& 336 & \textbf{1.128} & 0.873 & 1.134 & \textbf{0.857} & 0.505 & 0.487 & \textbf{0.495} & \textbf{0.484} & 1.168 & \textbf{0.880} & \textbf{1.147} & 0.884 \\
& 720 & 1.215 & 0.896 & \textbf{1.200} & \textbf{0.883} & \textbf{0.498} & \textbf{0.500} & 0.527 & 0.526 & \textbf{1.073} & \textbf{0.822} & 1.107 & 0.869 \\
\hline
ETTm1 & 24 & 0.323 & 0.369 & \textbf{0.291} & \textbf{0.367} & \textbf{0.383} & \textbf{0.403} & 0.404 & 0.429 & 0.316 & 0.371 & \textbf{0.312} & \textbf{0.371} \\ 
& 48 & 0.494 & 0.503 & \textbf{0.400} & \textbf{0.434} & \textbf{0.454} & \textbf{0.459} & 0.462 & 0.461 & 0.492 & 0.477 & \textbf{0.479} & \textbf{0.475} \\ 
& 96 & 0.678 & 0.614 & \textbf{0.485} & \textbf{0.470} & 0.481 & 0.463 & \textbf{0.470} & \textbf{0.459} & 0.619 & 0.581 & \textbf{0.566} & \textbf{0.528} \\ 
& 288 & 1.056 & 0.786 & \textbf{0.954} & \textbf{0.740} & 0.634 & 0.528 & \textbf{0.560} & \textbf{0.510} & 0.962 & 0.767 & \textbf{0.884} & \textbf{0.731} \\
& 672 & 1.192 & 0.926 & \textbf{1.016} & \textbf{0.792} & 0.606 & 0.542 & \textbf{0.546} & \textbf{0.519} & 1.168 & 0.837 & \textbf{0.897} & \textbf{0.720} \\
\hline
WTH & 96 & 0.552 & 0.535 & \textbf{0.548} & \textbf{0.526} & 0.266 & 0.366 & \textbf{0.263} & \textbf{0.336} & 0.450 & 0.466 & \textbf{0.350} & \textbf{0.407} \\ 
& 192 & 0.616 & 0.584 & \textbf{0.601} & \textbf{0.551} & 0.307 & 0.367 & \textbf{0.306} & \textbf{0.366} & 0.587 & 0.543 & \textbf{0.583} & \textbf{0.543} \\ 
& 336 & 0.702 & 0.620 & \textbf{0.599} & \textbf{0.562} & 0.359 & 0.395 & \textbf{0.358} & \textbf{0.395} & 0.648 & 0.576 & \textbf{0.635} & \textbf{0.549} \\
& 720 & 0.831 & 0.731 & \textbf{0.610} & \textbf{0.581} & \textbf{0.419} & \textbf{0.428} & 0.467 & 0.461 & 0.932 & 0.708 & \textbf{0.711} & \textbf{0.586} \\
\hline 
\end{tabular}}
\label{tab:result_multivariate}
\end{table}  

As shown in Table~\ref{tab:result_multivariate}, results of 34 out of 42 trials from our proposed tensor-augmented models surpass baselines, suggesting a significant improvement in model performance. For all dataset, the TEAFormer models have the better performance for the majority of the window sizes. Our model performs the best on WTH dataset. We argue that it is due to the WTH dataset has far more features and weather data has a more regular and predictable pattern than ETT. It has already been mentioned in many previous studies that ETT data has severe distribution shift challenge that has not been countered yet, which might impact TEAFormer's ability to forecast. We also notice that, the simpler the model structure, the better TEAFormer performs. Transformer is the start point of all attention-based time series forecasting studies, thus it has the simplest structure, followed by Informer, which is a modification to cope with multi-dimensional time series tasks. Autoformer is the latest and most complicated model among all three baselines, and we can observe that the baseline of Autoformer produces more better results than its corresponding TEA-counterparts than the other two models.

\subsection{Ablation Study}
\noindent{\bf{TEA-module in decoder}} In this study, we aim to address the current architecture challenge mentioned in Section~\ref{tea_transformer}, i.e., whether conducting attention operation on the core tensors throughout the entire model is a feasible solution. Here we decompose $\cX^{dec}$ and compute self-attention on the core tensor $\cC^{dec}$ to test our hypothesis.
\begin{equation}
\begin{aligned}
\cX^{dec} &= \hat{\cC}^{dec}\times_1 \bU_1\times_2\bU_2\times_3\bU_3,\\
\hat{\cC}^{dec} &= LayerNorm(\cC^{dec}+MSA(\cC^{dec},\cC^{dec},\cC^{dec}, mask)),
\end{aligned}
\end{equation}
Since we are only testing the effect of decomposing $\cX^{dec}$ during self-attention computation in this approach, we will not be using $\cC^{dec}$ for the rest of the computation. Thus we have
\begin{equation}
\begin{aligned}
\hat{\cX}^{dec} &= \hat{\cC}^{dec}\times_1 \bU_1\times_2\bU_2\times_3\bU_3,\\
\hat{\cX}^{out} &= LayerNorm(\hat{\cX}^{dec}+MSA(\hat{\cX}^{dec},\hat{\cX}^{enc},\hat{\cX}^{enc}, mask))  ,
\end{aligned}
\end{equation}
where $\hat{\cX}^{out}$ is the output of the current decoder layer, $\hat{\cX}^{enc}$ is the input from the encoder layer. We used this new TEAFormer method in Informer on ETTh1, denoted as TEA-Informer-ENC-DEC.
\begin{table}[h]  
\caption{Ablation study on ETTh1 dataset.} 
\centering 
\begin{tabular}{c c c c c c c} 
\hline
SeqLen & \multicolumn{2}{c}{{\bf Informer}} & \multicolumn{2}{c}{{\bf TEA-Informer}} & \multicolumn{2}{c}{{\bf TEA-Informer-ENC-DEC}}
\\ [0.5ex]  
\hline
& MSE & MAE & MSE & MAE & MSE & MAE\\
\hline
24 & 0.577 & 0.549 & \textbf{0.500} & \textbf{0.517} & 0.840 & 0.679\\
48 & 0.685 & 0.625 & \textbf{0.589} & \textbf{0.578} & 0.882 & 0.704\\
168 & \textbf{0.931} & \textbf{0.752} & 0.968 & 0.775 & 0.983 & 0.753\\ 
336 & \textbf{1.128} & 0.873 & 1.134 & \textbf{0.857} & 1.224 & 0.882\\
720 & 1.215 & 0.896 & \textbf{1.200} & \textbf{0.883} & 1.477 & 0.972\\
\hline 
\end{tabular}
\label{tab:ablation1}
\end{table}  
As shown in Table~\ref{tab:ablation1}, TEA-Informer-ENC-DEC does not perform well across all window sizes on ETTh1 dataset. On the contrary, it even performs worse than the baseline model. This result indicates that conducting attention operation on the core tensors throughout the entire model is not a feasible solution. We have two assumptions for why TEA-module cannot be implemented in decoder. First of all, as mentioned in Section~\ref{tea_transformer}, $\cX^{dec}$ is essential for forecasting, and we argue that the reduced dimension of $\cC^{dec}$ may not be sufficient to retain as much information as $\cX^{dec}$. Moreover, the factor matrices $\bU_1, \bU_2, \bU_3$ are not part of the learning process, thus it would not be accurate if we directly recombine the original $\bU_1,\bU_2,\bU_3$ that have not learned anything with the trained core tensor $\cC^{dec}$.

Apart from that, one key difference between the MSA modules in encoder and decoder is that attention mask is used in decoder. This attention mask is used to prevent the model from attending to certain positions in the input sequence, which is particularly important in time-series models, where the model should not have access to future tokens when predicting the current token. Theoretically, as a more abstract latent representation of the input tokens, it would be more difficult to learn with mask for a core tensor, and the prediction results would have more variance.

\noindent{\bf{TEA-Crossformer:}} Tensor expansion is a necessary step in input embedding as it allows the Transformer-based model to learn more key information. However, the larger expansion, the more computationally expensive the model becomes. In this part of ablation study, we want to test whether TEAFormer's usage of tensor compression techniques can reduce model's reliance on tensor size expansion. Crossformer~\cite{zhang2022crossformer} employs a segment-wise embedding, and manipulates data extensively using a two-stage attention architecture, in which attention is computed first along the time dimension then the feature dimension, such that hidden dimension is involved more frequently than our other baseline models, making it an ideal candidate for this part of study. It also implements a router mechanism that uses a fixed number of learnable matrix as intermediary to receive and send attention to reduce runtime complexity caused by computing attention across the batch. In this work, we only implement TEA-module in the cross-time stage where data is divided into segments along time dimension. We reduce the model's hidden dimension from 256 to 4 and obtain the following results.

\begin{table}[h]  
\caption{Ablation study results on ETTh1 and ILI datasets.} 
\centering 
\begin{tabular}{c c c c c c} 
\hline\hline   
Dataset & SeqLen & \multicolumn{2}{c}{Crossformer} & \multicolumn{2}{c}{TEA-Crossformer}
\\ [0.5ex]  
\hline
& & MSE & MAE & MSE & MAE\\
\hline\hline
ETTh1 & 24 & \textbf{0.797} & \textbf{0.632} & 0.862 & 0.668\\
& 48 & \textbf{0.720} & \textbf{0.581} & 1.063 & 0.774 \\
& 168 & 1.509 & 0.958 & \textbf{1.496} & \textbf{0.954} \\ 
& 336 & 1.414 & 0.936 & \textbf{1.366} & \textbf{0.873} \\
& 720 & \textbf{1.410} & \textbf{0.915} & 1.694 & 1.031 \\
\hline 
ILI & 24 & 7.064 & 1.903 & \textbf{6.572} & \textbf{1.844}\\
& 36 & 6.979 & 1.890 & \textbf{6.871} & \textbf{1.882}\\
& 48 & 7.189 & 1.930 & \textbf{6.833} & \textbf{1.530}\\
& 60 & 7.544 & 1.982 & \textbf{7.250} & \textbf{1.951}\\
\hline
\end{tabular}
\label{tab:ablation2}
\end{table}
\begin{figure*}[!htb]
\centering
\captionsetup{justification=centering}
\subfigure[MSE against sequence lengths]{\includegraphics[scale=0.38]{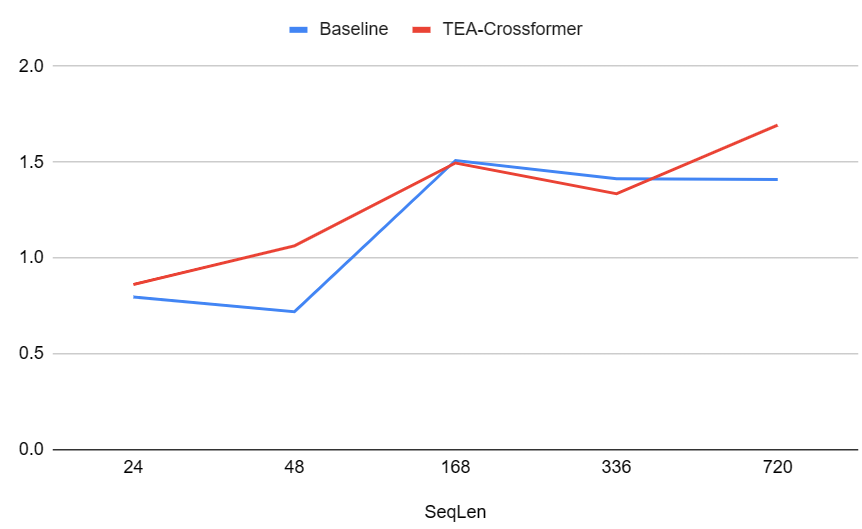}}\quad
\subfigure[MAE against sequence lengths]{\includegraphics[scale=0.38]{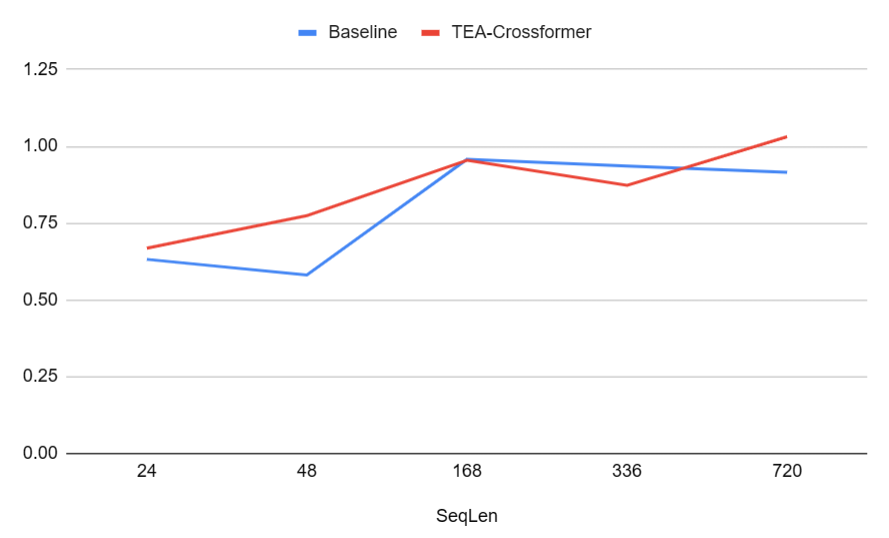}}
\caption{Performance of Crossformer vs TEA-Crossformer on ETTh1}
\label{fig:crossformer_performance}
\end{figure*}
According to Table~\ref{tab:ablation2}, TEA-Crossformer only surpass Crossformer on 2 out of 5 sequence lengths on ETTh1 dataset using the default hyperparameter settings. As shown in Figure~\ref{fig:crossformer_performance}, we are able to observe models' performance more intuitively: in cases where TEA-Crossformer surpassed Crossformer baseline, the differences are subtle. When facing long sequence forecasting, the Crossformer baseline has much less variance in both MSE and MAE than TEA-Crossformer. The errors in TEA-Crossformer increases proportionally to sequence length, but they tend to converge to a certain range in baseline Crossformer.

Additionally, we also test the models on national illness dataset (ILI\footnote{https://gis.cdc.gov/grasp/fluview/fluportaldashboard.html}) due to it has much fewer samples and shorter sequence length than other datasets. That is, in addition to the minimal hidden dimension, the learnable content is also relatively insufficient. On this dataset, as indicated in Table~\ref{tab:ablation2}, TEA-Crossformer outperforms the baseline on all sequence lengths, which illustrates the capability of TEA-based framework in learning rich information from small data. 

Our interpretation for the above results is that with the increase in model hidden dimension, the model learns more complicated patterns, especially for Crossformer which learns extensively. However, tensor compression, inherently, is a way of extracting the most predominant latent features in the original data, thus reduced the amount of information accessible to model learning. As a result, for a model that extensively manipulates data, only when the dataset does not provide enough data to learn from and tensor expansion is insufficient such that almost no latent patterns can be extracted, can tensor compression contribute.

\section{Conclusions and Future Work}\label{con_future}
We have proposed TEAFormer, a novel approach to multi-dimensional time series forecasting tasks that leverages tensor decomposition techniques in transformer through a Tensor-Augmented attention module, and deploys such module in time-series transformer models. The embedded data is treated as tensor mode and size expansion transformation, and is then tucker-decomposed into a core tensor and factor matrices. We use the core tensor as an aggregation of latent cross-dimensional and temporal representations, which contains valuable information about the interactions and contributions within the data. Multi-head self-attention is computed on core tensor such that self-attention's computational cost would be significantly reduced. By implementing TEAFormers with Transformer, Informer, and Autoformer, we successfully enhanced the performance of the three baseline models, proving Tensor Augmentation's effectiveness. 

We explored some potential directions for improvement and briefly discuss them: \textbf{1)} While computing MSA on core tensor indeed improves efficiency, extra computational cost is produced when decompose and re-construct the tensor objects. Our study has indicated that passing the core tensor solely throughout the transformer architecture is not a feasible solution due to dimension mismatch, insufficient information retained, as well as its incompatibility with the attention mask mechanism. An alternative way of implementing tensor-augmentaion needs to be found in order to ensure optimized efficiency. \textbf{2)} Our experiment of implementing TEAFormer with Crossformer suggests that tensor-augmentation module contributes significantly to learning with small data in short sequence forecasting tasks, given the minimal hidden dimension. This result demonstrates that TEAFormer has potential for performance enhancement when insufficient data or representation is given.

The potential of TEAFormer is not limited to our findings in this paper. The successful integration of tensor-augmented attention within established models underscores the versatility and adaptability of the TEAFormer methodology. This adaptability is particularly crucial in real-world applications where data characteristics and requirements can vary significantly. The ability of TEAFormer to maintain performance with reduced computational overhead highlights its practical value in resource-constrained environments, making advanced forecasting techniques more accessible and feasible across different industries.

The exploration of tensor-augmentation in handling small data and short sequence forecasting tasks further emphasizes its potential to revolutionize how models are trained and deployed, particularly in domains where data is limited or expensive to obtain. TEAFormer invites future research to delve deeper into the interplay between tensor decomposition and neural network architectures, potentially leading to more efficient and powerful models.

\clearpage
\bibliographystyle{plain}
\bibliography{NeurIPS_2023,tensor_network,tsf,cp}

\newpage
\setcounter{page}{1}
\appendix
\section*{\Large Appendices}


\section{Tensor Algebra} \label{sec:tensor}

\subsection{Tensor Low-Rank Structures} 
Consider an $M$-th order tensor $\mathcal{X}$ of dimension $D_1\times\cdots\times D_M$. 
If $\mathcal{X}$ assumes a (canonical) rank-$R$ 
{\em CP low-rank} structure, then it can be expressed as
\begin{equation} \label{eqn:cp}
    \mathcal{X} = \sum_{r=1}^R c_r \, \boldsymbol{u}_{1r} \circ \boldsymbol{u}_{2r} \circ \cdots \boldsymbol{u}_{Mr},
\end{equation}
where $\circ$ denotes the outer product, $\boldsymbol{u}_{mr}\in\mathbb{R}^{D_m}$ and $||\boldsymbol{u}_{mr}||_2 = 1$ for all mode $m \in [M]$ and latent dimension $r \in [R]$.
Concatenating all $R$ vectors corresponding to a mode $m$, we have
$\boldsymbol{U}_m = [\boldsymbol{u}_{m1}, \cdots, \boldsymbol{u}_{mR}] \in \mathbb{R}^{D_m \times R}$ which is referred to as the loading matrix for mode $m \in [M]$. 

If $\mathcal{X}$ assumes a rank-$(R_1, \cdots, R_M)$  
{\em Tucker low-rank} structure \eqref{eqn:tucker}, then it writes 
\begin{align*} 
\begin{split}
    \mathcal{X} = \mathcal{C}\times_1 \boldsymbol{U}_1\times_2\cdots\times_M \boldsymbol{U}_M = \sum_{r_1=1}^{R_1} \cdots \\
    \sum_{r_M=1}^{R_M} c_{r_1\cdots r_M} ( \boldsymbol{u}_{1 r_1} \circ \cdots \circ \boldsymbol{u}_{M r_M}),
\end{split}
\end{align*}
where $\boldsymbol{u}_{m r_m}$ are all $D_m$-dimensional vectors, and $c_{r_1\cdots r_M}$ are elements in the $R_1 \times \cdots \times R_D$-dimensional core tensor $\mathcal{C}$. 

{\em Tensor Train (TT) low-rank \citep{oseledets2011tensor}} approximates a $D_1\times\cdots\times D_M$ tensor $\mathcal{X}$ with a chain of products of third order {\it core tensors} $\mathcal{C}_i$, $i\in [M]$, of dimension $R_{i-1}\times D_i\times R_i$.  
Specifically, each element of tensor $\mathcal{X}$ can be written as 
\begin{equation} \label{eqn:tt}
    x_{i_1, \cdots, i_M} = \boldsymbol{c}_{1,1,i_1,:}^\top
    \times\boldsymbol{c}_{2,:,i_2,:}\times\cdots\times\boldsymbol{c}_{M,:,i_M,:}
    \times\boldsymbol{c}_{M+1,:,1,1}, 
\end{equation}
where $\boldsymbol{c}_{m,:,i_m,:}$ is an $R_{m-1}\times R_m$ matrix for $m\in[M]\cup\{M+1\}$.
The product of those matrices is a matrix of size $R_0 \times R_{M+1}$.
Letting $R_0 = 1$, the first core tensor $\mathcal{C}_1$ is of dimension $1 \times D_1 \times R_1$, which is actually a matrix and whose $i_1$-th slice of the middle dimension (i.e. $\boldsymbol{c}_{1,1,i_1,:}$) is actually a $R_1$ vector. 
To deal with the ``boundary condition'' at the end, we augmented the chain with an additional tensor $\mathcal{C}_{M+1}$ with $D_{M+1} = 1$ and $R_{M+1} = 1$ of dimension $R_M \times 1 \times 1$.  
So the last tensor can be treated as a vector of dimension $R_M$. 

CP low-rank \eqref{eqn:cp} is a special case where the core tensor $\mathcal{C}$ has the same dimensions over all modes, that is $R_m = R$ for all $m\in[M]$, and is super-diagonal. 
TT low-rank is a different kind of low-rank structure and it inherits advantages from both CP and Tucker decomposition.
Specifically, TT decomposition can compress tensors as significantly as CP decomposition, while its calculation is as stable as Tucker decomposition. 

\section{Experiment Results}
\subsection{Reproducibility}
The code to reproduce our experiment can be found at https://drive.google.com/drive/folders/1gzqWSQUCKOJh6c5\\
\_UtDK02V2JeMCrvUn?usp=sharing
\subsection{Resources}
All models are implemented in PyTorch and trained locally on a single NVIDIA GeForce RTX 3060 GPU and 16GB Memory. The runtime of one single experimental run varies depending on the input data, model and sequence length. 
\subsection{Supplemental performance charts}
\begin{figure*}[!htb]
  \centering
  \captionsetup{justification=centering}
  \subfigure[MSE against sequence lengths]{\includegraphics[scale=0.15]{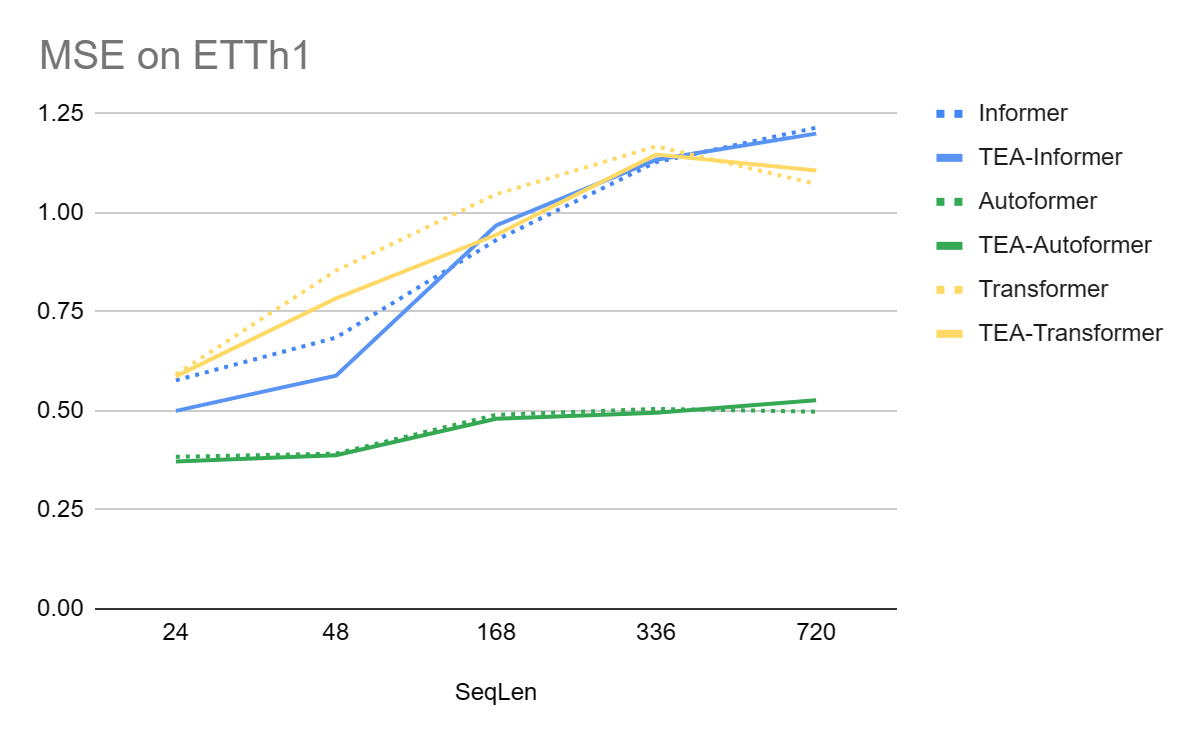}}\quad
  \subfigure[MAE against sequence lengths]{\includegraphics[scale=0.15]{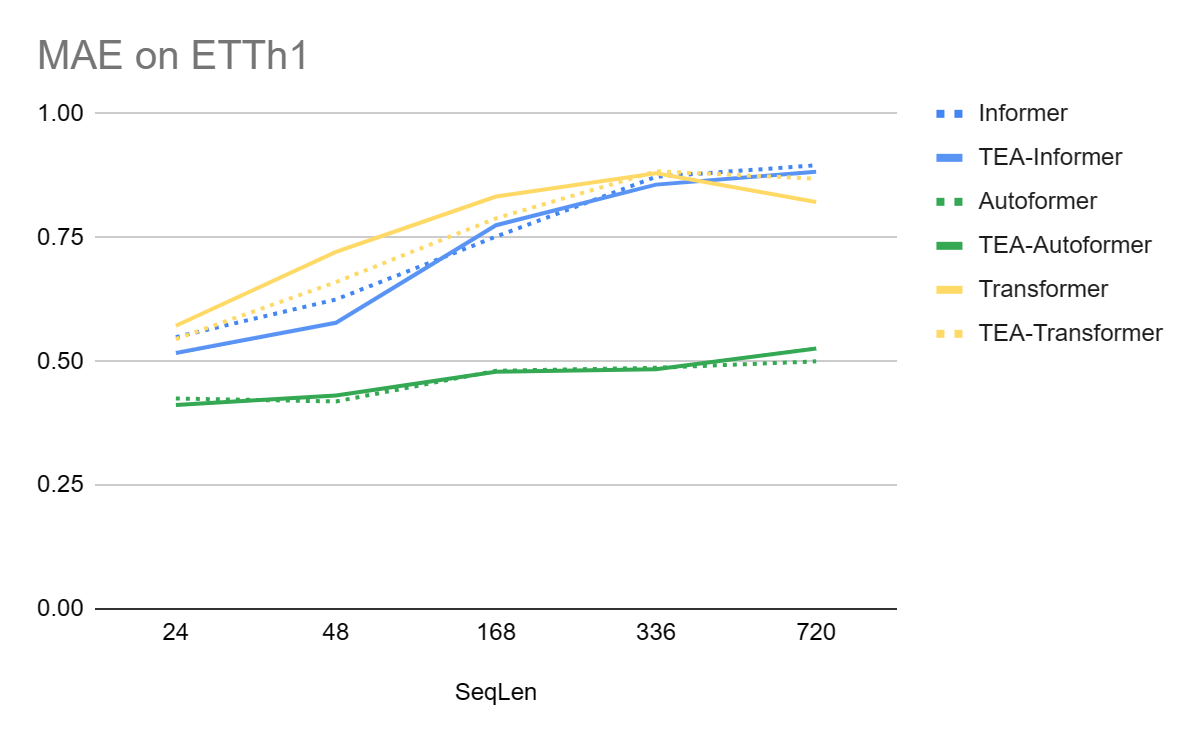}}
  \caption{Performance on ETTh1 }
  \label{fig:etth1_performance}
\end{figure*}
\begin{figure*}[!htb]
  \centering
  \captionsetup{justification=centering}
  \subfigure[MSE against sequence lengths]{\includegraphics[scale=0.15]{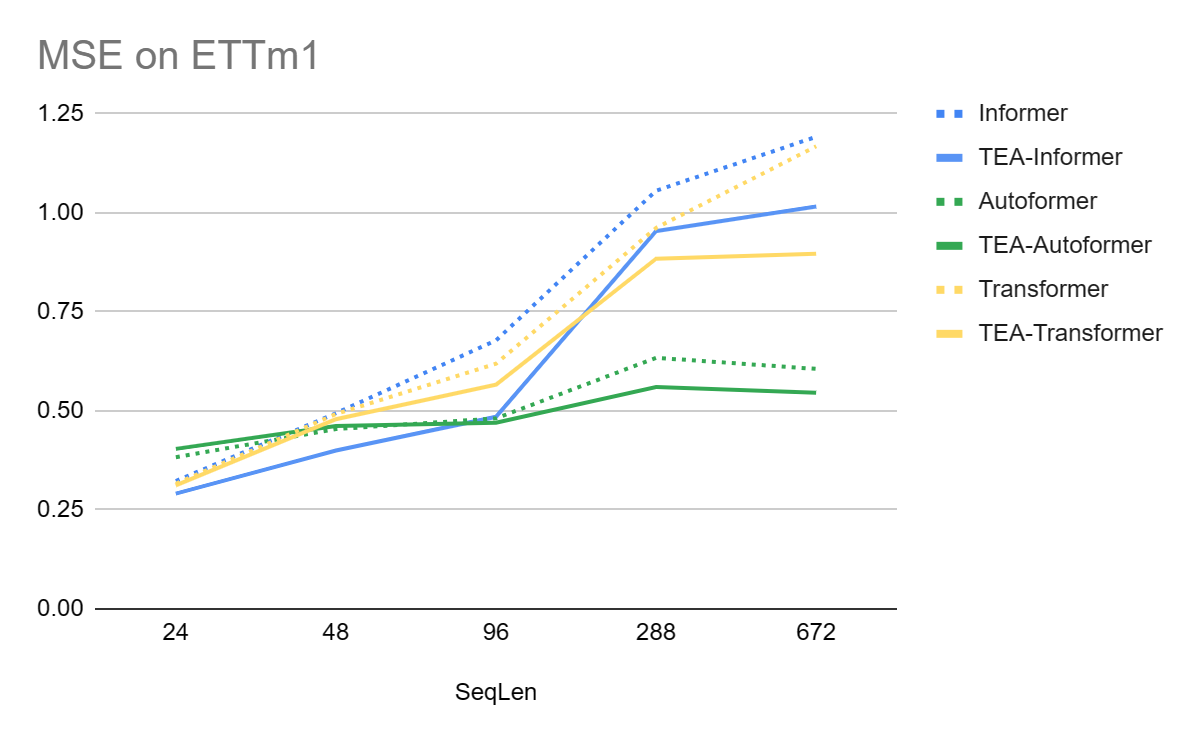}}\quad
  \subfigure[MAE against sequence lengths]{\includegraphics[scale=0.15]{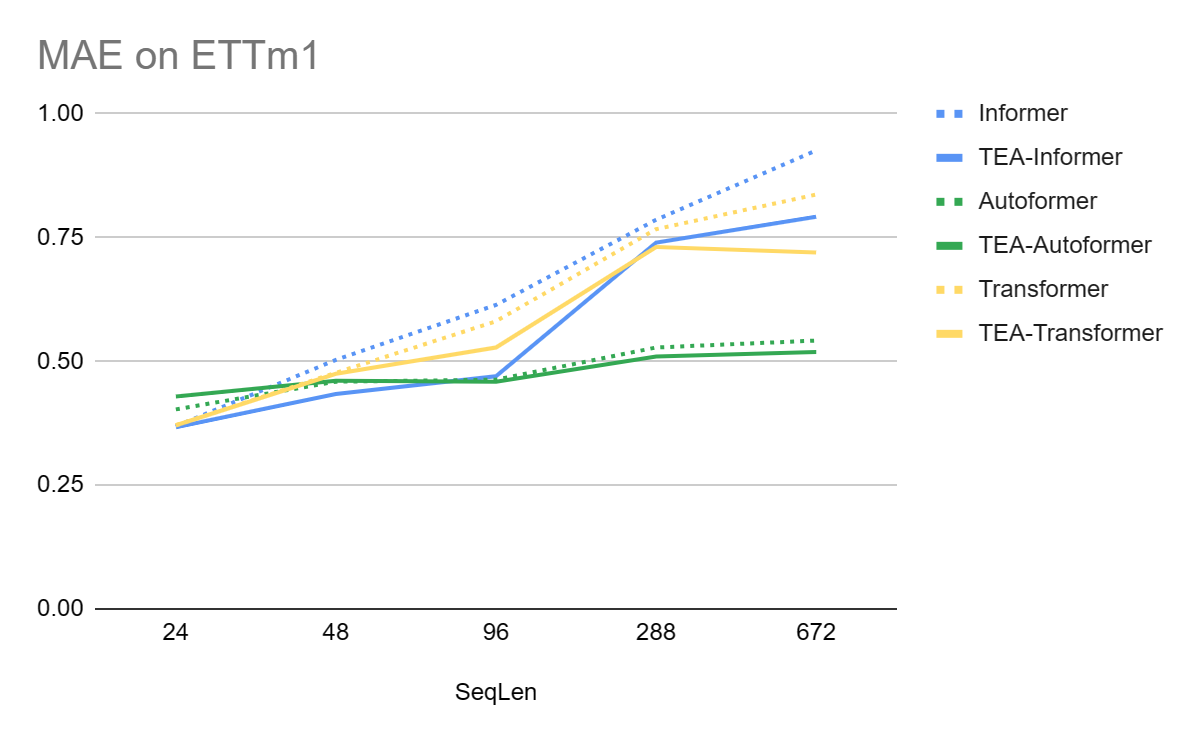}}
  \caption{Performance on ETTm1 }
  \label{fig:ettm1_performance}
\end{figure*}
\begin{figure*}[!htb]
  \centering
  \captionsetup{justification=centering}
  \subfigure[MSE against sequence lengths]{\includegraphics[scale=0.15]{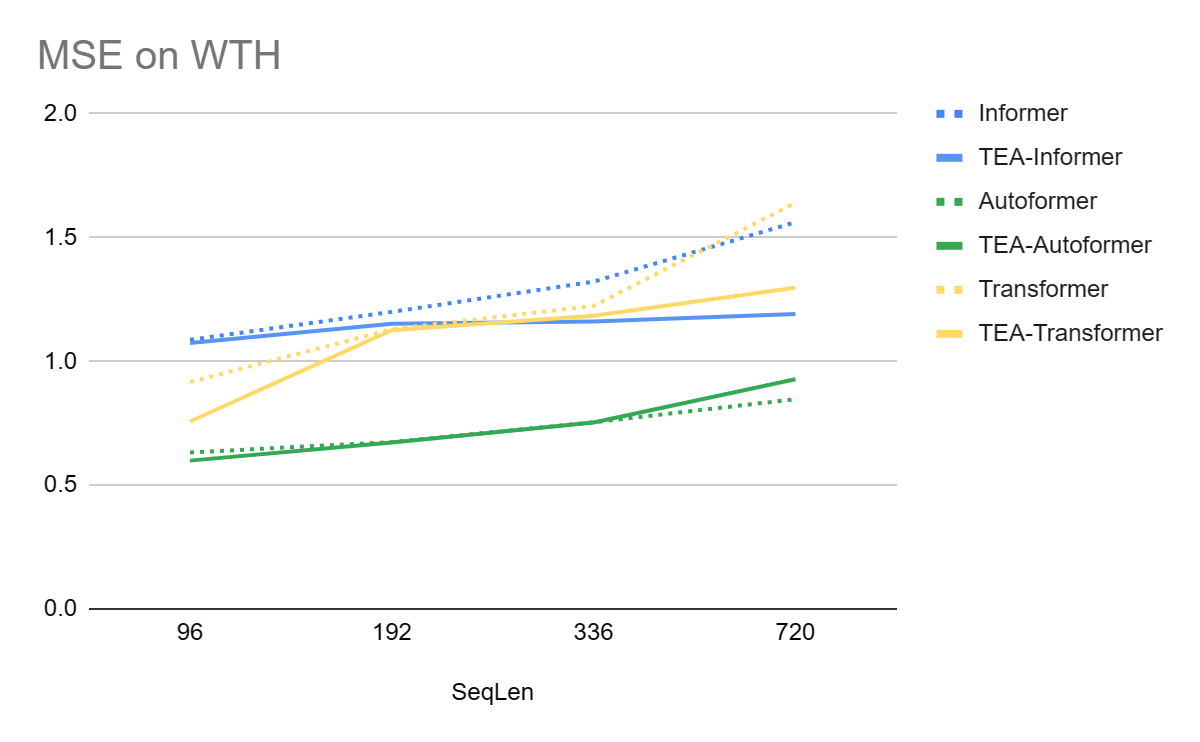}}\quad
  \subfigure[MAE against sequence lengths]{\includegraphics[scale=0.15]{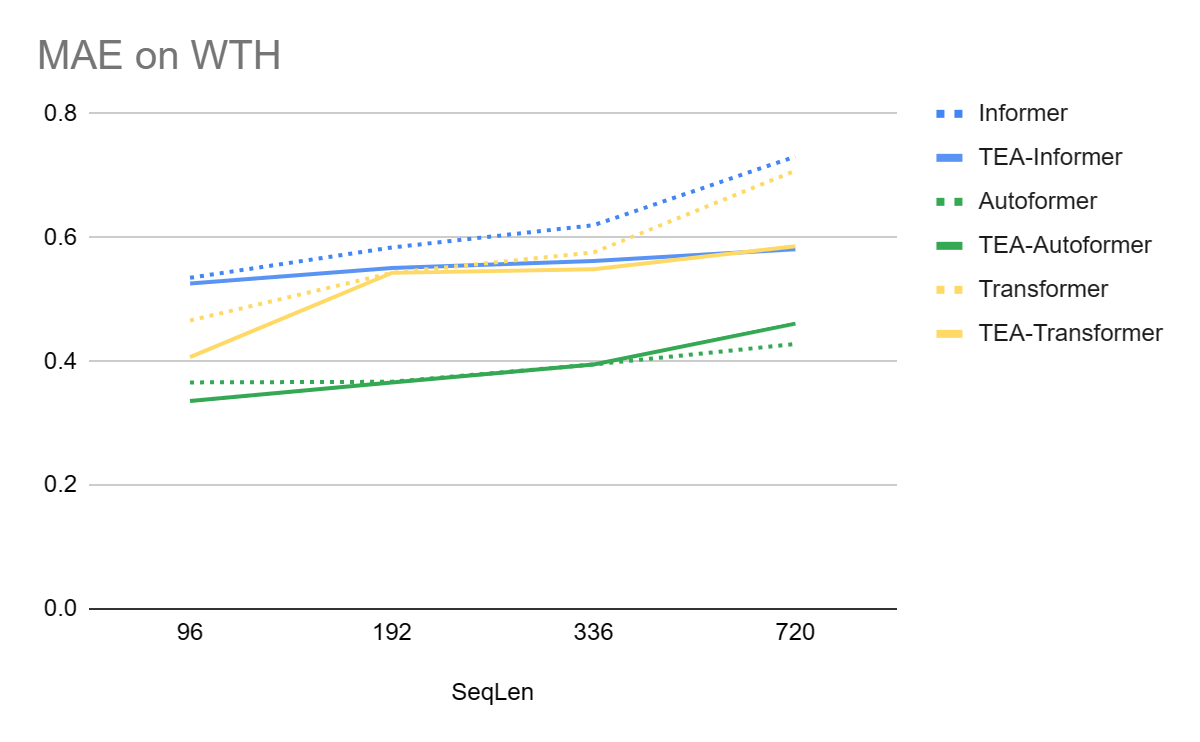}}
  \caption{Performance on WTH}
  \label{fig:wth_performance}
\end{figure*}

\end{document}